\documentclass[letterpaper, 10 pt, conference]{ieeeconf} 
\IEEEoverridecommandlockouts                          

\usepackage{graphics} 
\usepackage{epsfig} 
\usepackage{mathptmx} 
\usepackage{times} 
\usepackage{amsmath} 
\usepackage{amssymb}  
\usepackage[noadjust]{cite}
\usepackage{graphicx} 
\usepackage{booktabs} 
\usepackage{xcolor}
\usepackage{multirow}
\usepackage{adjustbox}
\usepackage{capt-of}
\usepackage{censor}
\usepackage{hyperref}
\usepackage[inkscapelatex=false]{svg}
\newtheorem{remark}{Remark}

\newcommand{\ms}[2]{#1{\tiny$\pm$#2}}

\title{\LARGE \bf
NoContactNoWorries: Estimating Contact through Vision and Proprioception for In-Hand Dexterous Manipulation
}

\author{
Soham Patil$^{1}$, \quad
Avirup Das$^{2}$, \quad
Sourabh Bhosale$^{1}$, \quad
Spandan Roy$^{1}$
\thanks{The work is partly supported by the UASAT project sponsored by MeitY, India.}
\thanks{$^{1}$The authors are with Robotics Research Center (RRC), International Institute of Information Technology (IIIT), Hyderabad, India, Emails: \texttt{soham.patil@research.iiit.ac.in, spandan.roy@iiit.ac.in}.}%
\thanks{$^{2}$Department of Computer Science, The University of Manchester, Email: \texttt{avirup.das@postgrad.manchester.ac.uk}.}%
}

\begin{document}
\twocolumn[{
\renewcommand\twocolumn[1][]{#1}
\maketitle
\thispagestyle{empty}
\vspace{-5mm} 
\begin{center}
    \centering
    \includegraphics[width=1.0\textwidth]{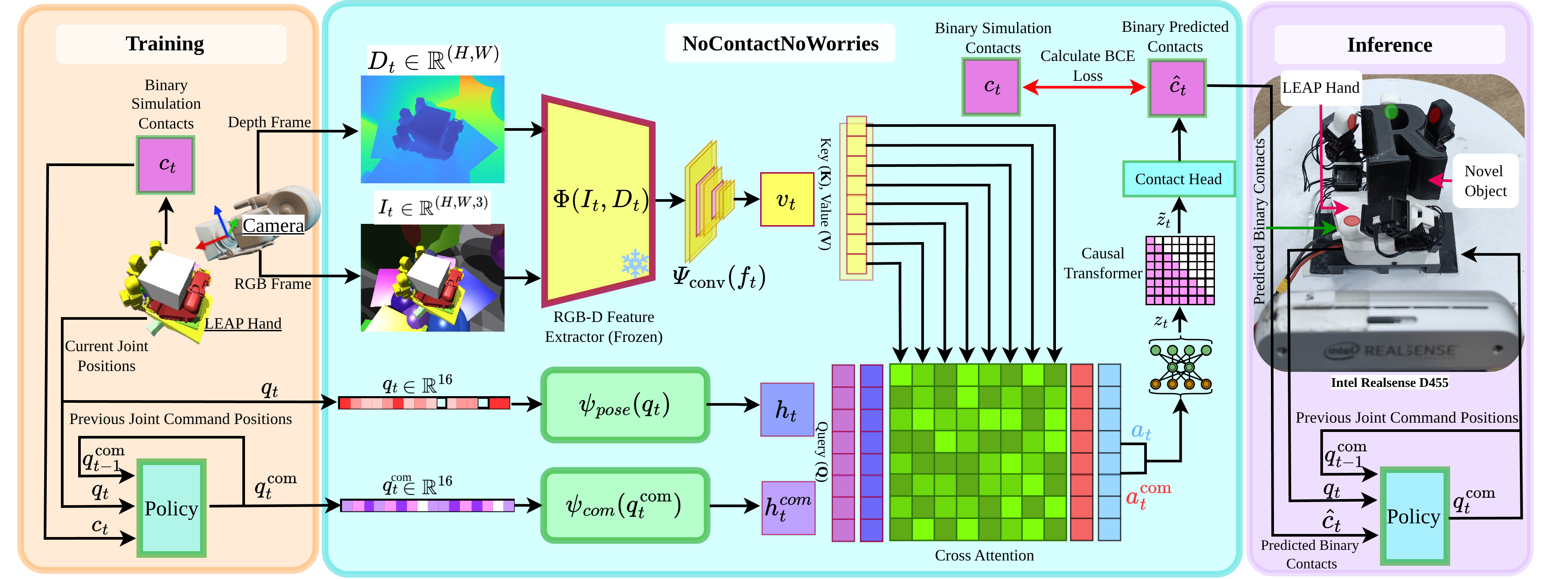}
    \captionof{figure}{\textbf{Training:} We collect vision, proprioception and contact data through a simulation environment by running a pretrained in-hand rotation policy. \textbf{NoContactNoWorries:} From synchronized RGB-D and proprioception at time $t$, the frozen encoder $\Phi(I_t, D_t)$ produces spatial features that are downsampled into visual tokens $v_t$. Current and commanded joint configurations ($q_t$, $q_t^{\text{com}}$) are embedded by $\psi_{\text{pose}}$ and $\psi_{\text{com}}$; these pose embeddings act as queries over $v_t$ via cross-attention, yielding a pose-conditioned visual representation. A causal transformer integrates these representations over a short window, and a contact head outputs the binary contact vector $\hat{c}_t$ for each fingertip site. The predicted contact can be fed to a manipulation policy as a surrogate tactile signal. \textbf{Inference:} Hardware Setup and its interaction with the Policy and inferred contact. }
    \vspace{-2mm}
    \label{fig:pipeline}
\end{center}
}]
\begin{abstract}
Perceiving physical contact is fundamental to dexterous manipulation. While robots often rely on dedicated hardware tactile sensors, humans exhibit a remarkable ability to infer contact by integrating visual information with an innate sense of their body's pose and movement. Inspired by this embodied perceptual skill, we investigate whether a robot can learn to infer contact from vision, an approach that also offers a scalable alternative to tactile hardware specifically for binary contact estimation, which faces practical challenges in cost, fragility, and integration. We present NoContactNoWorries, a transformer-based multimodal framework that fuses RGB-D vision with the robot’s proprioception to infer binary contact states as a pseudo-tactile signal for hand-object interactions. We validate by training a single contact prediction model on multiple objects and show that the inferred contact signal supports downstream reinforcement learning agents for in-hand object reorientation, generalizing to novel objects. Experiments in both simulation and on a real-world robot validate our approach, highlighting the feasibility of inferring contact from vision and proprioception. Project Page: \url{https://soham2560.github.io/no-contact-no-worries/}
\end{abstract}

\vspace{-1mm} 
\section{Introduction}
\label{sec:introduction}
Dexterous manipulation is a central challenge in robotics because it requires precise control of finger-object contacts under diverse and uncertain interactions~\cite{chen2022system}. Effective control depends on contact-rich feedback (e.g., force distribution, slip, stability, and interaction-conditioned object pose) that is often ambiguous or occluded in vision, but is directly observable through touch~\cite{yang2025anyrotate,yin2023rotating}. Vision and proprioception provide global scene context and internal state, yet tactile sensing uniquely exposes localized contact dynamics that underpin in-hand manipulation, grasp adjustment, and fine motor control~\cite{liuvtdexmanip,lee2020making}.
Despite this utility, tactile sensing has not matched vision in deployment scalability. Cameras are ubiquitous, but generalist dexterity in unstructured settings increasingly requires ego-centric perception (e.g., wrist-mounted RGB-D), which suffers from self-occlusion during finger-object interaction. By contrast, tactile hardware is often costly, fragile, and difficult to scale in coverage. High-fidelity sensors such as GelSight~\cite{yuan2017gelsight} and DIGIT~\cite{lambeta2020digit} measure rich contact signals but only over small patches, while large-area alternatives (e.g., pressure skins, resistive arrays) typically trade resolution for substantial integration overhead (wiring, calibration). This asymmetry motivates a practical question: \textit{can we infer contact from existing modalities (ego-centric vision and proprioception) when tactile sensing is unavailable or impractical?}

Biological evidence suggests that contact can be inferred from multimodal sensory cues. Humans integrate vision and haptics to estimate contact-related properties~\cite{ernst2002humans}, while the primary somatosensory cortex (S1) receives convergent visual and proprioceptive inputs~\cite{rizzolatti1988functional,de2005neuronal}. These findings motivate our hypothesis that contact can be predicted without dedicated tactile sensors using ego-centric vision and proprioception.

This work proposes pseudo-tactile sensing for dexterous in-hand manipulation, i.e., predicting binary fingertip contact from ego-centric RGB-D and proprioception without tactile measurements. We use a transformer that fuses RGB-D tokens with both current and commanded joint configurations via cross-attention, and captures temporal dynamics with causal attention, producing per-fingertip contact estimates at each frame (Fig.~\ref{fig:pipeline}). We evaluate predictive accuracy in simulation and on a real LEAP hand, including generalization to held-out objects, and show that the predicted contact can replace oracle/tactile contact for in-hand reorientation policies trained in simulation, enabling direct transfer to hardware without policy retraining. The predicted signal can be used directly by downstream controllers, or fused with tactile sensing when available.
\vspace{-1mm}
\section{Related Work}
\vspace{-1mm}
Vision is one of the most widely used sensing modalities in robotic manipulation due to its ubiquity and high information density. Vision-based methods have achieved strong performance in dexterous settings, ranging from learning policies from videos~\cite{chen2025vividex} to visuomotor controllers that reorient objects with unseen geometries~\cite{jain2019learning, chen2023visual, chi2023diffusion, ze20243ddiffusion}. However, purely visual observations often provide weak supervision over contact-dependent phenomena such as slip, force distribution, and local stability cues, especially under self-occlusion during finger-object interaction. Recent analysis~\cite{lu2026when} shows visual information can be under-utilized during motion-transition sub-phases that require target localization, contributing to inconsistent generalization across tasks.

Tactile feedback provides direct, localized evidence of contact, and has been shown to improve contact-rich skills from re-grasping to fine motor tasks~\cite{yin2025learning, zhao2025tactile}. For example, AnyRotate~\cite{yang2025anyrotate} achieves long-horizon, gravity-invariant object rotation with sim-to-real transfer by leveraging dense tactile signals for contact localization and slip-aware control. Despite these successes, tactile hardware remains difficult to deploy at scale due to cost, fragility, limited coverage, and integration overhead.

To combine complementary cues, prior work has fused vision and touch for state estimation and contact-aware control. Representative approaches include visuo-tactile occupancy or geometry-alignment for localizing surface contact~\cite{yuan2024robot, huang20253d}, and policy-level fusion of wrist RGB-D, proprioception, and fingertip tactile signals for robust in-hand reorientation~\cite{qi2023general}. In parallel, self-supervised multimodal pretraining has learned shared visuo-tactile representations using masked reconstruction and/or cross-modal prediction~\cite{sferrazza2024power, liu2024masked, heng2025vitacformer, chen2023visuo, cui2024interrep, george2025vital, liu2025vitamin}. While effective, these methods typically assume tactile inputs are available at deployment, or jointly retrain downstream policies to exploit the learned representations~\cite{george2025vital}.
It is noteworthy that \cite{upvital2025} also aims to bypass physical tactile sensors. However, their prediction model is used only to calculate training rewards, implying that the predictions are not used for any downstream tasks as a substitute for tactile feedback. As observed in~\cite{ng1999policy}, relying solely on reward shaping also risks unintentional changes in the goal of the task without actually giving the robot new sensory information. In contrast, our method acts as a ``virtual sensor": it predicts a pseudo-tactile signal from ego-centric vision and proprioception, and provides it directly to the robot as a real-time observation. This allows downstream controllers to actively react to contacts during execution, without requiring policy retraining, specialized hardware, or task-specific reward engineering.


\section{Methodology}
\label{sec:methodology}
We address the problem of estimating binary contact signals at a fixed set of fingertip locations using visual and proprioceptive sensing, in the absence of physical tactile measurements. Formally, at each timestep $t$, the robot observes an RGB image $I_t\in\mathbb{R}^{H\times W\times 3}$, a depth map $D_t\in\mathbb{R}^{H\times W}$, the current joint configuration $q_t\in\mathbb{R}^{J}$, and a commanded joint configuration $q_t^{\text{com}}\in\mathbb{R}^{J}$ -- modalities that are commonly accessible on most dexterous robotic platforms. The task is to predict a binary contact vector $\hat{c}_t \in \{0,1\}^K$, where each dimension corresponds to a tactile sensing site. These sensing modalities individually provide incomplete access to contact measurements. Visual inputs offer indirect cues (local shading, deformation, occlusion patterns, etc.) that correlate with contact, but are often ambiguous and viewpoint-dependent. Depth observations provide explicit hand-object geometry and help disambiguate proximity from occlusion, but lack semantic richness. Proprioceptive inputs do not directly encode external object geometry. However, under closed-loop manipulation they can correlate with contact events through control intent and contact-induced motion deviations. Critically, many of these cues manifest only through changes over time. As such, effective contact estimation requires fusing these complementary but individually insufficient signals and reasoning over both their spatial structure and temporal evolution. Our approach integrates these inputs over short temporal windows, enabling contact prediction through learned representations that capture dynamic correlations between hand configuration, observed geometry, and motion intent.

\subsection{RGB-D Feature Extraction}
To estimate contact from vision, we employ an encoder that  maps RGB-D inputs into a shared spatial feature space:
\begin{align*}
    f_t = \Phi(I_t, D_t), \quad  f_t \in \mathbb{R}^{C \times H' \times W'}.
\end{align*}

Keeping $\Phi$ frozen allows us to isolate and study spatiotemporal reasoning across modalities without entangling it with low-level visual representation learning. Training a reliable RGB-D encoder from scratch would require large-scale, modality-specific supervision. We therefore adapt $\Phi$ from an existing RGB-D semantic segmentation backbone~\cite{du2024asym}, which processes the two modalities through asymmetric streams: a convolutional network for RGB and a transformer-based module for depth. Local spatial attention and cross-modal fusion layers combine appearance and geometry, making the representation particularly well suited to our task, where visual cues must be grounded in spatial context (e.g., finger-object distance) to resolve contact-relevant features.

To adapt the frozen visual features for contact prediction, we append a lightweight convolutional module $\Psi_{\text{conv}}$, which is trained jointly with the rest of the model. This module performs two stages of strided convolutions, reducing spatial resolution while projecting features to a fixed embedding dimension $D$ to produce a set of compact spatial tokens:
\begin{align*}
    v_t = \Psi_{\text{conv}}(f_t) \in \mathbb{R}^{N \times D},
\end{align*}
where $N$ is the number of spatial tokens. These tokens form the intermediate visual representation used in subsequent stages of the task.

\subsection{Pose-Conditioned Spatial Fusion}

To resolve the ambiguities inherent in vision-based contact estimation, we incorporate proprioceptive state information into the visual stream through a cross-attention mechanism conditioned on both the current and commanded joint configurations. This treats proprioception as a dynamic selector over the visual scene, allowing the robot to extract context-sensitive features relevant to its configuration and control objectives.

Given the visual tokens $v_t \in \mathbb{R}^{N \times D}$ obtained from RGB-D features, we embed the current joint configuration $q_t \in \mathbb{R}^J$ and the commanded configuration $q_t^{\text{com}} \in \mathbb{R}^J$ (which represents the target joint positions provided to the low-level proportional-derivative (PD) controller) using learned linear projections $\psi$ and $\psi_{\text{com}}$:
\begin{align*}
    h_t = \psi(q_t), \quad h_t^{\text{com}} = \psi_{\text{com}}(q_t^{\text{com}}); \quad h_t, h_t^{\text{com}} \in \mathbb{R}^D.
\end{align*}

These embedded proprioceptive vectors serve as distinct queries passed through a shared multi-head attention over $v_t$, producing two attended visual representations:
\begin{align*}
    a_t = \text{Attn}(h_t, v_t, v_t), \
    a_t^{\text{com}} = \text{Attn}(h_t^{\text{com}}, v_t, v_t); \
    a_t, a_t^{\text{com}} \in \mathbb{R}^D.
\end{align*}

This pose-conditioned attention mechanism is designed to reflect the complementary roles of reactive and anticipatory sensing. The current pose $q_t$ captures the realized state of the hand and allows the model to identify visual features that align with current finger configurations (e.g., unexpected surface contact, deformation, or slip). In contrast, the commanded pose $q_t^{\text{com}}$ encodes the control intent and enables the model to focus on regions that are likely to be relevant for upcoming contact events. This design is informed by control-theoretic principles, where feedback and feedforward signals jointly inform action, and finds strong parallels in biological systems: the motor cortex and posterior parietal cortex integrate proprioceptive and visual information through both actual and intended movement representations to support sensorimotor control.

While both queries attend to the same visual tokens, their semantic and functional roles are distinct and asymmetric. Treating them symmetrically would obscure this distinction and limit expressivity. Hence, to produce a fused, proprioception-aware visual embedding that is dynamically conditioned on both internal state and task intent, we concatenate the resulting features and pass them through a learned multilayer perceptron:
\begin{align*}
    z_t = \text{MLP}([a_t; a_t^{\text{com}}]) \in \mathbb{R}^D.
\end{align*}

\subsection{Temporal Modeling for Contact Estimation}

Contact is often indistinguishable from mere proximity in a single frame. Instead, it emerges through temporal cues like trajectory deviations, persistent occlusions, or cumulative changes in hand-object geometry.

To address this, we form a sequence of per-frame pose-conditioned visual representations $\{z_{t-T+1}, \ldots, z_t\}$ over a fixed-length temporal window $T$, and model their evolution using a causal Transformer encoder:
\begin{align*}
    \tilde{z}_{t} = \text{Transformer}_{\text{causal}}(z_{t-T+1}, \ldots, z_t),
\end{align*}
where $\tilde{z}_t \in \mathbb{R}^D$ is the contextualized representation at the current timestep. The final token is passed through a linear contact head to predict the binary contact vector:
\begin{align*}
    \hat{c}_t = \sigma(W_c \tilde{z}_t + b_c) \in [0,1]^K,
\end{align*}
where $\sigma$ denotes element-wise sigmoid activation. This temporal formulation allows the model to accumulate evidence across time and infer contact from dynamic cues that may be ambiguous or imperceptible from static input alone.

\begin{figure}[h!]
    \centering
    \includegraphics[width=0.48\textwidth]{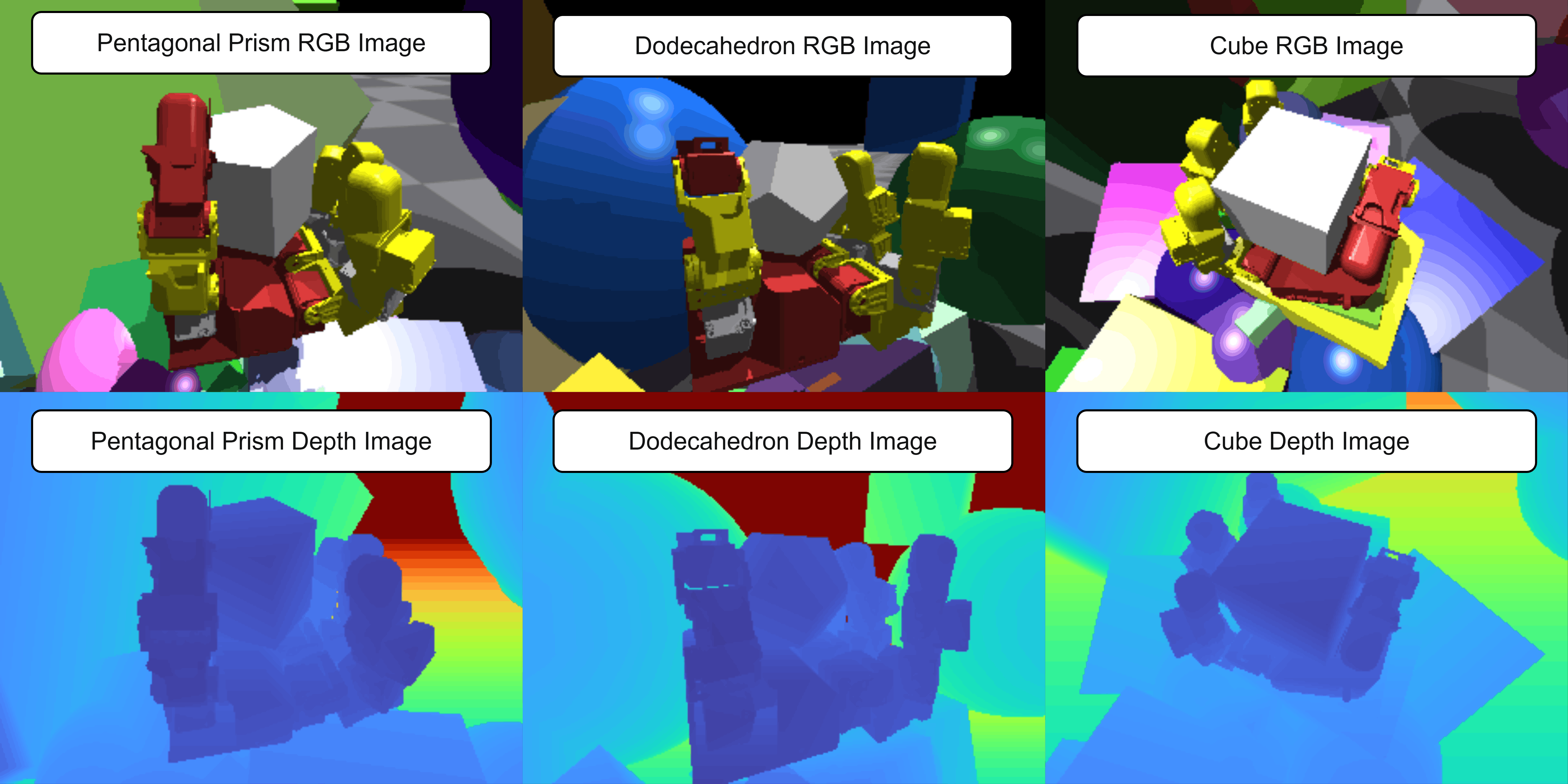}
    \caption{Data augmentation in simulation with randomized background objects, visual adjustments in color, contrast, brightness, and reflectivity.}
    \vspace{-4mm}
    \label{fig:image_augmentations}
\end{figure}
\section{System Setup and Training Protocol}

We collect contact-labeled demonstrations in simulation using the LEAP Hand platform~\cite{shaw2023leap}, built on NVIDIA Isaac Gym. Visual feedback is provided by a wrist-mounted Intel RealSense D455 camera, positioned beyond its minimum depth range. While multi-view setups resolve occlusion, they restrict agents to calibrated workcells~\cite{huang2023earl}. We retain the wrist-mounted constraint to evaluate our method on self-contained hands, as recent benchmarks show it is sufficient if occlusions are handled~\cite{koczy2025learning}. We capture synchronized RGB-D images at 30 Hz, with the entire system controlled from a workstation equipped with an NVIDIA RTX 4090 GPU. The hand executes in-hand object rotation tasks with a Gated Recurrent Unit (GRU)-based policy trained using Proximal Policy Optimization (PPO), with training episodes initialized from a cache of stable grasps~\cite{qi2023hand}. Key hyperparameters include $\gamma=0.99$, $\lambda=0.95$, and a learning rate of $3 \times 10^{-4}$. To guide the learning process, we adopt the reward structure from Shaw et al.~\cite{shaw2023leap}, which combines a primary reward for the object's angular velocity with penalties for deviating from a stable pose, excessive motor effort, and high object velocity to prevent drops. During RL training in simulation, the policy is provided with ground-truth contacts from the physics engine to learn an effective contact-aware strategy.

Ground-truth contact labels are obtained from the simulator's physics engine (PhysX), which provides binary contact signals at four fingertip locations ($K=4$). These sensing sites were selected to correspond to the center of the distal phalanges of the four active fingers on the LEAP hand, mirroring the feasible physical mounting points for the real-world FSR sensors used during hardware evaluation. To support sim-to-real transfer and improve policy robustness, we apply targeted domain randomization to both dynamics and perception. Physics parameters such as object mass (0.01-0.25\, kg), friction coefficients (0.3-3.0), PD controller gains, and object scale are randomized across episodes. To handle perceptual variability, we inject visual distractors, randomize camera viewpoints, add Gaussian sensor noise, and apply mild perturbation forces on the object. Example frames illustrating these augmentations are shown in Fig.~\ref{fig:image_augmentations}.

\begin{figure}[!h]
    \centering
    \includegraphics[width=1.0\linewidth]{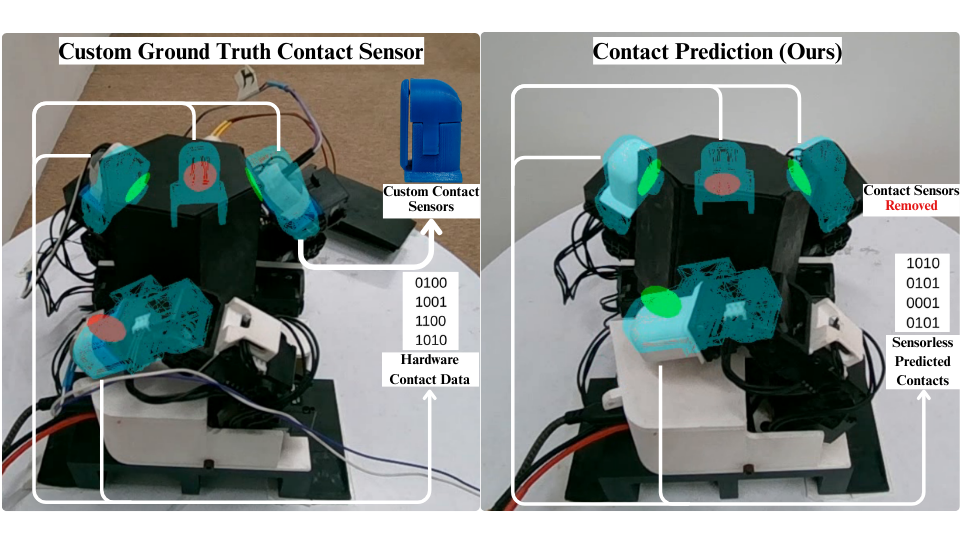}
    \caption{\textbf{Real-World Evaluation Setup.} (\textit{Left}) Ground-truth contact labels acquired using $<0.5$\,mm force-sensitive resistors (FSRs) attached to the LEAP Hand fingertips, limiting geometric discrepancy with the simulation model. (\textit{Right}) During model inference and policy deployment, the physical tactile sensors are removed. The model infers the corresponding binary contact states (indicated by colored overlays, green indicating contact) using only vision and proprioception.}
    \vspace{-5mm}
    \label{fig:sensor_comparison}
\end{figure}

To obtain ground-truth contact labels in the real world, custom, low-profile resistive force sensors were attached to the LEAP Hand fingertips (Fig.~\ref{fig:sensor_comparison}, left). Force signals are binarized using a threshold of $0.1$\,N, a standard heuristic aligned with our simulation binary labels. These physical sensors are used strictly for evaluation metrics (F1 scores) while being disconnected from the inference pipeline and are completely withheld from the downstream policy during deployment, which relies exclusively on the model's predicted pseudo-tactile signals (Fig.~\ref{fig:sensor_comparison}, right).

\begin{figure}[!h]
    \centering
     \includegraphics[width=\columnwidth]{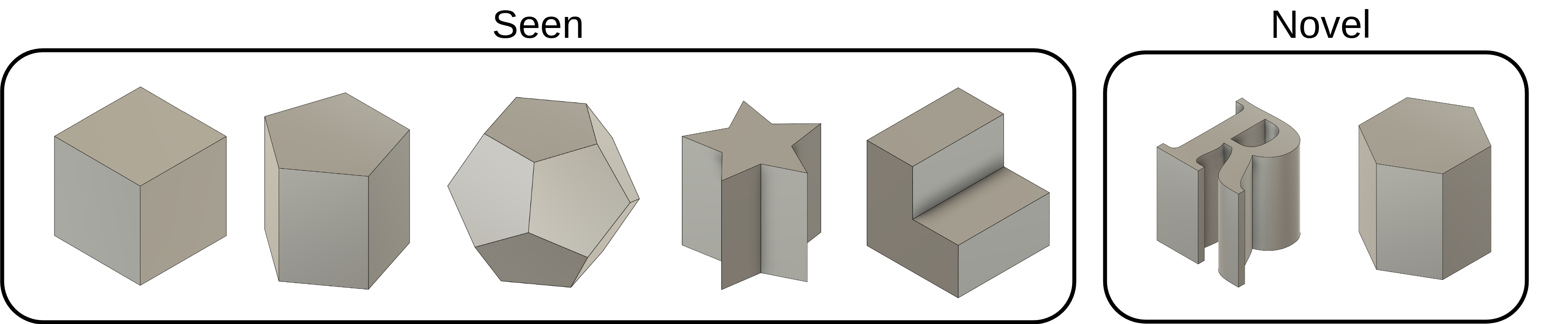}
    \caption{\textbf{Object Sets for Experiments.} (a) Five primitive objects seen during training: cuboid, pentagonal prism, extruded star, dodecahedron and stairs. (b) Novel objects: an extruded letter `R' and a hexagonal prism held out from all training, used to evaluate zero-shot generalization.
    \vspace{-4mm}
    }
    \label{fig:object_sets}
\end{figure}

The dataset comprises rollouts collected in simulation using the five training object geometries (Fig.~\ref{fig:object_sets}). We record 50 trajectories of 15\,s each (10 per object), where each trajectory is generated by an independently trained policy (distinct seeds, converged checkpoints and state-action occupancy) to avoid controller-specific leakage. Predictive accuracy metrics are reported over 10 object-stratified, policy-disjoint random splits, i.e., in each split, we hold out 2 trajectories per object for validation and train on the remaining 8 per object, ensuring that all validation trajectories come from policies unseen during training.

The contact prediction model is trained using supervised learning over this dataset. Each timestep provides synchronized RGB ($240{\times}320{\times}3$), depth ($240{\times}320$), joint state ($\mathbb{R}^{16}$), commanded joint state ($\mathbb{R}^{16}$), and ground-truth contact labels ($\mathbb{R}^{4}$). The model is optimized (excluding the frozen RGB-D encoder) with a binary cross-entropy loss over all timesteps $T$:
\begin{equation*}
    \mathcal{L}_{\text{contact}} = - \frac{1}{TK}\sum_{t} \sum_{k=1}^{K} \left[ c_t^{(k)} \log \hat{c}_t^{(k)} + (1 - c_t^{(k)}) \log (1 - \hat{c}_t^{(k)}) \right].
\end{equation*}

The model is implemented in PyTorch and trained using AdamW with a constant learning rate of $1\times10^{-4}$, weight decay $0.01$ and effective batch size $128$. Both the causal transformer and cross-attention modules use a single attention layer with one head and embedding dimension 256, operating over a temporal window $T=8$.

\section{Experiments}
\label{sec:experiments}
We evaluate our proposed method for predicting binary contact signals from RGB-D and proprioceptive inputs in the absence of physical tactile sensing. Our experiments are structured to answer three core questions: (i) How accurate is the predicted contact signal across both seen and novel objects in simulation and the real world? (ii) What is the contribution of each architectural component to the overall predictive performance? (iii) Can predicted contact signals effectively substitute for true tactile input in downstream control tasks on real robotic hardware?

\begin{table*}[t]
\centering
\caption{Contact prediction F1 scores on simulated and real trajectories for seen and novel objects.
\vspace{-1mm}
}
\label{tab:combined_f1}
\begin{adjustbox}{max width=\textwidth}
\begin{tabular}{l|l|ccccc|cc}
\toprule
\multirow{2}{*}{\textbf{Setting}} & \multirow{2}{*}{\textbf{Model Variant}} & 
\multicolumn{5}{c|}{\textbf{Seen Objects}} & \multicolumn{2}{c}{\textbf{Novel Objects}} \\
 & & Cuboid & Pentagonal Prism & Dodecahedron & Star & Stairs & Hexagonal Prism & Letter R \\
\midrule
\multirow{9}{*}{\textbf{Simulated}} 
  & (a) Pose-only (query-asymmetry) & 0.81 ± 0.026 & 0.80 ± 0.024 & 0.79 ± 0.017 & 0.74 ± 0.025 & 0.75 ± 0.013 & 0.78 ± 0.018 & 0.73 ± 0.022 \\
  & (b) Pose-only (symmetric) & 0.78 ± 0.014 & 0.75 ± 0.022 & 0.77 ± 0.014 & 0.72 ± 0.020 & 0.72 ± 0.012 & 0.74 ± 0.033 & 0.69 ± 0.031 \\
  & (c) Pose-only (MLP) & 0.74 ± 0.023 & 0.71 ± 0.012 & 0.74 ± 0.013 & 0.65 ± 0.020 & 0.65 ± 0.025 & 0.69 ± 0.024 & 0.60 ± 0.031 \\
  & (d) Pose-delta only & 0.72 ± 0.027 & 0.71 ± 0.024 & 0.71 ± 0.033 & 0.61 ± 0.031 & 0.62 ± 0.020 & 0.67 ± 0.055 & 0.57 ± 0.023 \\
  & (e) Vision-only & 0.66 ± 0.030 & 0.64 ± 0.018 & 0.65 ± 0.031 & 0.57 ± 0.046 & 0.58 ± 0.037 & 0.62 ± 0.042 & 0.55 ± 0.045 \\
  & (f) No temporal modeling & 0.78 ± 0.015 & 0.81 ± 0.023 & 0.75 ± 0.025 & 0.78 ± 0.020 & 0.77 ± 0.026 & 0.78 ± 0.024 & 0.68 ± 0.031 \\
  & (g) Kinematic depth baseline & 0.69 ± 0.020 & 0.67 ± 0.018 & 0.68 ± 0.021 & 0.61 ± 0.024 & 0.63 ± 0.022 & 0.65 ± 0.019 & 0.59 ± 0.021 \\
  & \textcolor{black}{Full (mismatched command)} & \textcolor{black}{0.91 ± 0.012} & \textcolor{black}{0.87 ± 0.018} & \textcolor{black}{0.86 ± 0.020} & \textcolor{black}{0.83 ± 0.015} & \textcolor{black}{0.86 ± 0.028} & \textcolor{black}{0.86 ± 0.021} & \textcolor{black}{0.82 ± 0.028} \\
  & \textbf{Full (Proposed)} & \textbf{0.93 ± 0.010} & \textbf{0.90 ± 0.013} & \textbf{0.91 ± 0.015} & \textbf{0.88 ± 0.012} & \textbf{0.88 ± 0.025} & \textbf{0.89 ± 0.026} & \textbf{0.87 ± 0.026} \\
\midrule
\multirow{9}{*}{\textbf{Real}} 
  & (a) Pose-only (query-asymmetry) & 0.72 ± 0.029 & 0.71 ± 0.034 & 0.70 ± 0.017 & 0.63 ± 0.024 & 0.68 ± 0.023 & 0.69 ± 0.027 & 0.63 ± 0.016 \\
  & (b) Pose-only (symmetric) & 0.70 ± 0.027 & 0.69 ± 0.023 & 0.68 ± 0.015 & 0.61 ± 0.031 & 0.66 ± 0.020 & 0.67 ± 0.027 & 0.61 ± 0.028 \\
  & (c) Pose-only (MLP) & 0.66 ± 0.026 & 0.65 ± 0.031 & 0.64 ± 0.032 & 0.57 ± 0.017 & 0.61 ± 0.030 & 0.63 ± 0.034 & 0.58 ± 0.028 \\
  & (d) Pose-delta only & 0.52 ± 0.023 & 0.50 ± 0.026 & 0.51 ± 0.031 & 0.43 ± 0.028 & 0.49 ± 0.024 & 0.50 ± 0.029 & 0.41 ± 0.019 \\
  & (e) Vision-only & 0.48 ± 0.028 & 0.45 ± 0.026 & 0.46 ± 0.032 & 0.34 ± 0.018 & 0.44 ± 0.026 & 0.45 ± 0.022 & 0.36 ± 0.018 \\
  & (f) No temporal modeling & 0.69 ± 0.024 & 0.68 ± 0.034 & 0.67 ± 0.035 & 0.65 ± 0.034 & 0.65 ± 0.015 & 0.66 ± 0.024 & 0.62 ± 0.021 \\
  & (g) Kinematic depth baseline & 0.55 ± 0.024 & 0.52 ± 0.021 & 0.53 ± 0.027 & 0.42 ± 0.019 & 0.50 ± 0.023 & 0.51 ± 0.022 & 0.45 ± 0.018 \\
  & \textcolor{black}{Full (mismatched command)} & \textcolor{black}{0.83 ± 0.027} & \textcolor{black}{0.80 ± 0.035} & \textcolor{black}{0.80 ± 0.020} & \textcolor{black}{0.68 ± 0.018} & \textcolor{black}{0.77 ± 0.025} & \textcolor{black}{0.74 ± 0.036} & \textcolor{black}{0.69 ± 0.032} \\
  & \textbf{Full (Proposed)} & \textbf{0.84 ± 0.026} & \textbf{0.83 ± 0.033} & \textbf{0.82 ± 0.016} & \textbf{0.71 ± 0.031} & \textbf{0.79 ± 0.025} & \textbf{0.80 ± 0.027} & \textbf{0.74 ± 0.029} \\
\bottomrule
\end{tabular}
\end{adjustbox}
\vspace{-4mm}
\end{table*}

\subsection{Ablations and Baselines}
To isolate the role of each sensing modality and architectural component, we evaluate several controlled variants of the \textbf{Full(Proposed)} model described in Section~\ref{sec:methodology}. In pose-only ablations, the visual token set $v_t$ is removed entirely. The fusion module then operates solely on embedded pose tokens by applying self-attention over the pair $[h_t, h_t^{\text{com}}]$ at each timestep, after which the resulting representation is processed by the same temporal encoder and contact head as the full model.

\textbf{(a) Pose-only w/ attention (query-asymmetry)}: Vision free variant that applies two-token self-attention over the per-timestep pair $[h_t, h_t^{\text{com}}]$, then preserves token identity by concatenating the attended current and command tokens before the fusion MLP, mirroring the `two-stream` pre-temporal topology of the full model while excluding visual information.

\textbf{(b) Pose-only w/ attention (symmetric)}: Same two-token self-attention over $[h_t, h_t^{\text{com}}]$, but removes directional asymmetry by mean-pooling the two attended tokens before fusion, testing whether the asymmetry of separating current vs. command streams is important beyond attention alone.

\textbf{(c) Pose-only w/o attention}: Replaces the cross-attention with a simple MLP over the concatenated pose embeddings, retaining the temporal model and contact head. This serves as a baseline for assessing whether temporal modeling over pose geometry alone can solve the task without any attention-based fusion.

\textbf{(d) Pose-delta only}: Provides only the difference vector $q_t^{\text{com}} - q_t$ as input, removing both vision and absolute pose information. This tests the sufficiency of motion intent alone, without explicit modeling of current configuration or interactions between pose signals.

\textbf{(e) Vision-only}: Removes all proprioceptive inputs and uses only RGB-D tokens $v_t$ from the same frozen backbone used in the full model. To obtain per-contact-site features without pose-conditioned queries, $K$ learned contact-site query embeddings are used (one per fingertip/contact site) that cross-attend to $v_t$ at each timestep, producing site-specific features that are fed to the same causal temporal transformer and contact head. This quantifies how much contact predictability can be recovered from visual observations alone under the same modeling assumptions as the proposed method (i.e., without explicit kinematic projection of contact sites into the image).

\textbf{(f) No temporal modeling}: A single-frame model using both RGB-D and proprioception, but without any temporal sequence modeling. This tests the hypothesis that instantaneous observations suffice and provides contrast to the full model, which reasons over temporal evolution.

\textbf{(g) Kinematic depth}: A geometric baseline that predicts contact using forward kinematics and depth proximity. At each timestep, fingertip positions are obtained via forward kinematics in the hand base frame and transformed into the wrist-camera frame using calibrated extrinsics. The depth image is back-projected (using camera intrinsics) into a local point cloud around the projected fingertip pixel, and contact is predicted for fingertip when the minimum Euclidean distance between fingertip position and the observed surface points falls below a threshold (obtained by temporal median filtering to reduce flicker).

\subsection{Contact Prediction Performance}

We begin by evaluating the predictive quality of our model in estimating binary contact signals from vision and proprioception. These experiments aim to assess (i) the overall accuracy of the predicted contact signal, (ii) the model’s ability to generalize to novel object geometries, and (iii) the necessity and contribution of individual architectural components. We adopt the \emph{F1 score} as our primary evaluation metric, computed per-timestep and averaged across contact sites. This metric captures the trade-off between precision (avoiding false positives) and recall (detecting true contacts). We report results across four settings: simulated trajectories on seen objects, simulated trajectories on novel objects, real-world replicas of training objects, and real-world replicas of novel objects. Each result is averaged over 10 splits. Detailed per-object F1 scores are provided in Table~\ref{tab:combined_f1}.

\begin{figure}[!h]
    \centering
    \includegraphics[width=\columnwidth]{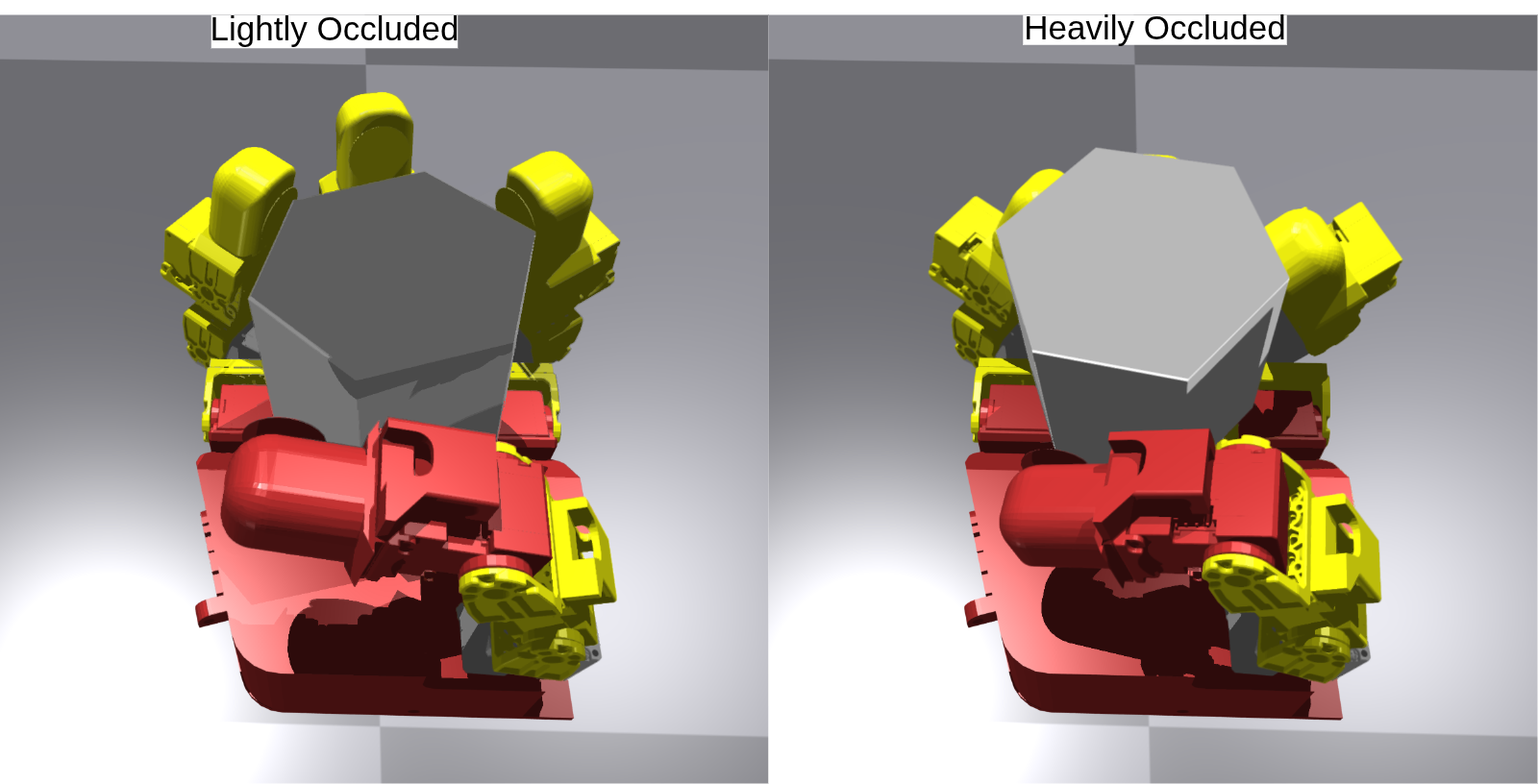}
    \caption{\textbf{Visual Occlusion during In-Hand Manipulation.} Simulated views from the wrist-mounted camera during interaction with the Hexagonal Prism. (\textit{Left}) A lightly occluded configuration where fingertip contact regions are largely visible. (\textit{Right}) A heavily occluded configuration where the object geometry obstructs the view of the distal phalanges. As quantified in Table~\ref{tab:visibility}, vision-only variants degrade in such frames.}
    \vspace{-7mm}
    \label{fig:occlusion_examples}
\end{figure}

Across all settings, the full model achieves the highest performance, confirming that combining RGB-D vision with proprioceptive state and motion intent enables accurate contact prediction without tactile sensors. In simulation, performance exceeds 0.9 F1 on convex objects and remains strong on challenging non-convex geometries. Real-world performance degrades due to sensing noise and calibration mismatch but remains robust (0.82-0.84 on convex objects; 0.71 star, 0.74 R), indicating effective sim-to-real transfer.

\paragraph{Visibility and geometric ambiguity} To isolate the effect of visual occlusion (see Fig.~\ref{fig:occlusion_examples}), we conduct a controlled analysis in simulation using the Hexagonal Prism where camera geometry and mesh visibility are known exactly. A contact site is labeled \emph{visible} if both fingertip and local object surface lie within the camera frustum and are not self-occluded under ray casting; otherwise it is labeled \emph{occluded}. Across trajectories, contact regions are occluded in approximately 61\% of frames (averaged over contact sites). Table~\ref{tab:visibility} reports F1 conditioned on visibility.

\begin{table}[!h]
\vspace{-2mm}
\caption{F1 score conditioned on contact visibility.}
\centering
\label{tab:visibility}
\begin{tabular}{lcc}
\toprule
\textbf{Model} & \textbf{Visible} & \textbf{Occluded} \\
\midrule
Kinematic depth baseline & 0.81 & 0.54 \\
Vision-only & 0.79 & 0.51 \\
Pose-only (best) & 0.75 & 0.81 \\
Full (Proposed) & \textbf{0.93} & \textbf{0.85} \\
\bottomrule
\end{tabular}
\vspace{-2mm}
\end{table}

Vision-only performance improves when contact sites are visible (0.79) but drops sharply under occlusion (0.51), confirming that vision alone provides incomplete information about fingertip-object interaction. The kinematic depth baseline performs similarly to vision-only and likewise degrades under occlusion. This baseline has direct access to calibrated geometry yet remains substantially below the full model, indicating that the transformer learns predictive structure beyond instantaneous geometric proximity, including motion-conditioned contact dynamics and temporal consistency. In contrast, the full multimodal model remains robust across visibility regimes (0.93 visible, 0.85 occluded) by leveraging proprioceptive cues when visual evidence is incomplete.

\begin{figure}[!h]
    \centering
    \vspace{-2mm}
    \includegraphics[width=1.0\linewidth]{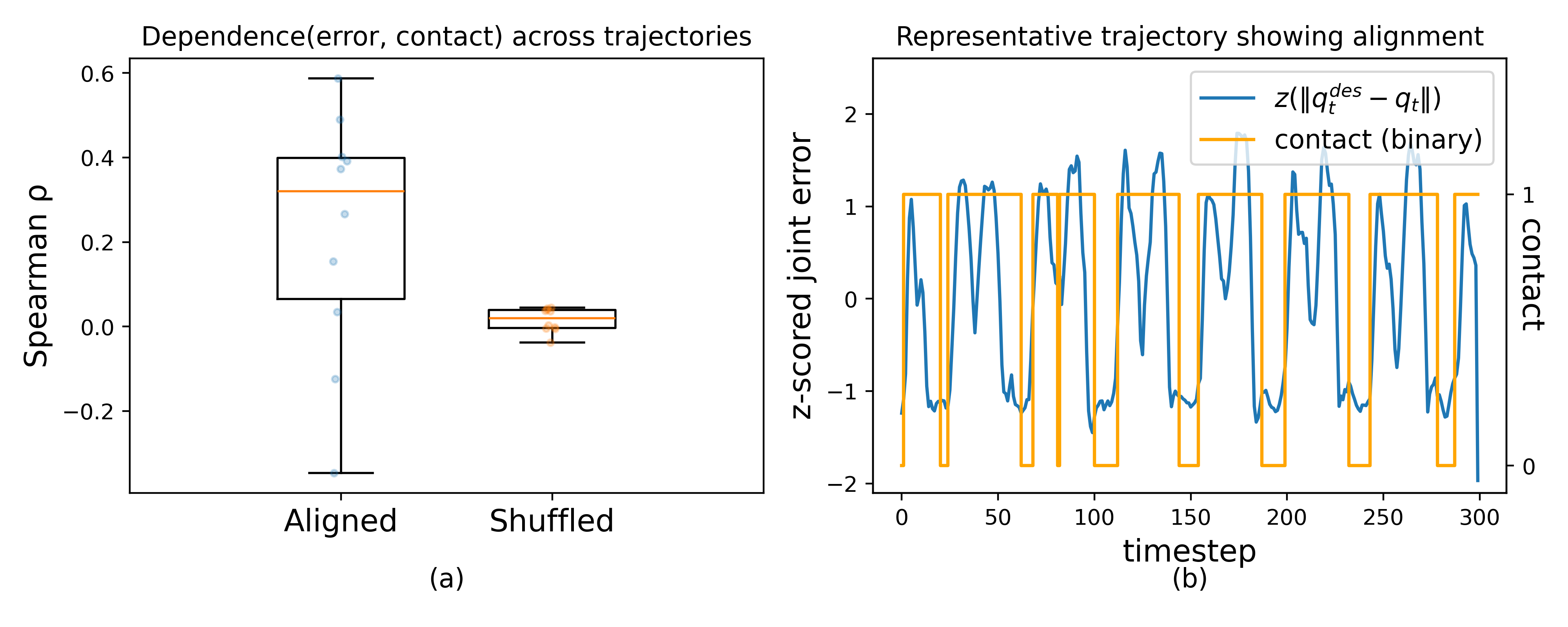}
    \vspace{-4mm}
    \caption{Statistical dependence between joint tracking error and contact events. (a) Distribution of Spearman rank correlation between $||q_t^{\text{com}}-q_t||$ and contact events. The time-shuffled baseline removes temporal alignment and yields near-zero dependence. (b) Representative trajectory showing alignment between tracking error and contact transitions.}
    \vspace{-4mm}
    \label{fig:pose_contact_correlation}
\end{figure}

\paragraph{Why pose-only can compete with vision-only}
All trajectories are generated by closed-loop policies, so $q_t^{\text{com}}$ encodes control intent, and deviations between realized and commanded motion correlate with contact onset and release. Figure~\ref{fig:pose_contact_correlation} shows that configuration error magnitude exhibits statistically significant monotonic dependence with contact labels, while temporal shuffling removes this dependence. Multimodal fusion remains necessary because visual geometry and proprioceptive intent provide complementary information, and the full model consistently outperforms all single-modality variants.

\paragraph{Mismatched-command stress test}
Since $q_t^{\text{com}}$ encodes motion intent, it could in principle provide shortcut cues that correlate with contact timing. To test whether the full model over-relies on this token, we perform a \emph{mismatched-command} evaluation in which $q_t^{\text{com}}$ is replaced by a command token sampled from a different object (and, for novel objects, from a different trajectory) while keeping RGB-D and current proprioception unchanged. As shown in Table~\ref{tab:combined_f1}, F1 decreases only slightly across both seen and novel objects in simulation and on hardware. This indicates that our model is not simply memorizing an intent$\rightarrow$contact prior, but instead grounds contact inference in multimodal cues and their temporal consistency, using $q_t^{\text{com}}$ primarily as auxiliary context rather than a dominant shortcut signal.

Among pose-only variants, the query-asymmetry design outperforms symmetric and MLP baselines, indicating the importance of preserving directional structure between current and commanded configurations. The pose-delta baseline performs comparably to vision-only in real-world tests, suggesting that discrepancies between intended and realized motion contain useful but incomplete contact cues. Temporal modeling further improves performance and helps disambiguate true contact from transient occlusion or near-contact events by leveraging short-horizon motion consistency. Removing the causal encoder reduces F1 across objects, particularly for non-convex geometries where instantaneous proximity is ambiguous.

Overall, while performance degrades under real sensing noise, the full model consistently outperforms all baselines, demonstrating that multimodal fusion with temporal reasoning provides a robust tactile-free contact signal suitable for downstream manipulation.

\subsection{Downstream Performance}

\begin{table*}
\centering
\caption{Ablation analysis of the system (CRA and CRR). Results are averaged over 10 trials. Bold marks the best non-oracle result in each column. The oracle uses ground-truth contact and is shown as an upper-bound reference. }
\vspace{-3mm}
\label{tab:ablation_results}
\begin{adjustbox}{max width=\textwidth}
\setlength\tabcolsep{1.5pt}
\small
\begin{tabular}{lcccccccccccccc}
\toprule
Method 
& \multicolumn{2}{c}{\textbf{Cuboid}} 
& \multicolumn{2}{c}{\textbf{Pentagonal Prism}} 
& \multicolumn{2}{c}{\textbf{Dodecahedron}} 
& \multicolumn{2}{c}{\textbf{Star}} 
& \multicolumn{2}{c}{\textbf{Stairs}} 
& \multicolumn{2}{c}{\textbf{Hexagonal Prism}} 
& \multicolumn{2}{c}{\textbf{Letter R}} \\
\cline{2-3} \cline{4-5} \cline{6-7} \cline{8-9} \cline{10-11} \cline{12-13} \cline{14-15}
& CRR & CRA 
& CRR & CRA 
& CRR & CRA 
& CRR & CRA 
& CRR & CRA 
& CRR & CRA 
& CRR & CRA \\
\hline
(a) Pose-only(query-asym) 
 & \ms{101.97}{11.02} & \ms{15.83}{0.75} 
 & \ms{106.47}{5.85} & \ms{17.71}{3.26} 
 & \ms{89.36}{16.03} & \ms{15.95}{0.71} 
 & \ms{80.64}{7.11} & \ms{11.32}{2.46} 
 & \ms{80.05}{18.67} & \ms{12.12}{2.53} 
 & \ms{86.62}{16.47} & \ms{11.73}{1.15} 
 & \ms{76.67}{3.19} & \ms{12.01}{1.39} \\
(b) Pose-only(symmetric) 
 & \ms{94.49}{7.28} & \ms{14.51}{1.82} 
 & \ms{113.31}{16.92} & \ms{20.03}{2.12} 
 & \ms{100.64}{16.19} & \ms{15.90}{3.07} 
 & \ms{81.16}{8.77} & \ms{13.45}{2.57} 
 & \ms{83.46}{7.39} & \ms{14.48}{3.14} 
 & \ms{106.47}{8.19} & \ms{18.50}{1.52} 
 & \ms{68.20}{5.07} & \ms{13.54}{0.55} \\
(c) Pose-only(MLP)
 & \ms{108.37}{9.71} & \ms{17.22}{1.75} 
 & \ms{101.39}{11.43} & \ms{16.11}{2.68} 
 & \ms{99.42}{11.27} & \ms{17.42}{1.56} 
 & \ms{89.60}{19.96} & \ms{\textbf{15.61}}{3.23} 
 & \ms{76.49}{11.06} & \ms{11.95}{0.59} 
 & \ms{111.50}{14.34} & \ms{15.32}{3.08} 
 & \ms{68.46}{3.74} & \ms{9.49}{0.86} \\
(d) Pose-delta only
 & \ms{103.32}{8.49} & \ms{17.02}{1.57} 
 & \ms{84.95}{5.15} & \ms{15.09}{0.43} 
 & \ms{88.54}{9.08} & \ms{13.97}{2.73} 
 & \ms{86.98}{8.28} & \ms{14.34}{1.70} 
 & \ms{73.47}{6.51} & \ms{13.09}{2.90} 
 & \ms{94.29}{10.17} & \ms{13.50}{1.36} 
 & \ms{83.24}{12.95} & \ms{13.06}{1.87} \\
(e) Vision-only
 & \ms{85.88}{8.64} & \ms{15.70}{1.73} 
 & \ms{89.70}{9.10} & \ms{14.73}{1.85} 
 & \ms{85.16}{9.37} & \ms{11.62}{0.64}
 & \ms{88.14}{7.07} & \ms{13.28}{1.03} 
 & \ms{94.88}{5.49} & \ms{15.01}{2.50} 
 & \ms{106.12}{11.18} & \ms{17.68}{2.61} 
 & \ms{76.10}{7.15} & \ms{14.96}{0.76} \\
(f) No temporal modeling
 & \ms{84.34}{9.89} & \ms{13.03}{0.74} 
 & \ms{103.79}{9.41} & \ms{17.09}{1.32} 
 & \ms{109.38}{10.71} & \ms{16.50}{1.37} & \ms{71.14}{8.90} & \ms{12.95}{2.51} & \ms{87.31}{4.02} & \ms{14.65}{0.64} & \ms{109.51}{11.88} & \ms{15.73}{3.31} & \ms{86.32}{15.68} & \ms{12.76}{1.24} \\
\textbf{Proprioception Only (No Contact)} 
 & \ms{113.09}{12.24} & \ms{19.46}{2.81} 
 & \ms{111.84}{15.58} & \ms{19.36}{2.46} 
 & \ms{113.38}{17.45} & \ms{18.95}{4.51} 
 & \ms{90.01}{18.10} & \ms{13.25}{3.34} 
 & \ms{98.02}{15.48} & \ms{\textbf{17.57}}{3.92} 
 & \ms{117.03}{19.00} & \ms{\textbf{19.68}}{3.27} 
 & \ms{89.72}{15.98} & \ms{\textbf{15.29}}{3.21} \\
\hline
\textbf{with Contact (Oracle)} 
 & \ms{124.43}{5.89} & \ms{19.10}{1.06} 
 & \ms{124.65}{13.70} & \ms{19.56}{0.92} 
 & \ms{121.34}{6.81} & \ms{18.43}{0.62} 
 & \ms{97.98}{8.19} & \ms{14.95}{1.19} 
 & \ms{107.19}{7.95} & \ms{16.88}{1.57} 
 & \ms{124.18}{7.34} & \ms{18.34}{1.28} 
 & \ms{99.11}{19.88} & \ms{15.39}{1.37} \\
\textbf{Ours (Proposed)} 
 & \ms{\textbf{122.99}}{11.93} & \ms{\textbf{20.24}}{1.46} 
 & \ms{\textbf{116.67}}{16.35} & \ms{\textbf{20.65}}{2.48} 
 & \ms{\textbf{118.77}}{8.49} & \ms{\textbf{19.40}}{2.05} 
 & \ms{\textbf{92.77}}{11.79} & \ms{14.37}{1.83} 
 & \ms{\textbf{107.00}}{10.22} & \ms{16.83}{2.08} 
 & \ms{\textbf{118.92}}{13.31} & \ms{17.64}{2.12} 
 & \ms{\textbf{94.80}}{18.72} & \ms{15.12}{1.76} \\
\bottomrule
\end{tabular}
\end{adjustbox}
\vspace{-3mm}
\end{table*}

\begin{figure}[!h]
    \centering
    \includegraphics[width=0.85\linewidth]{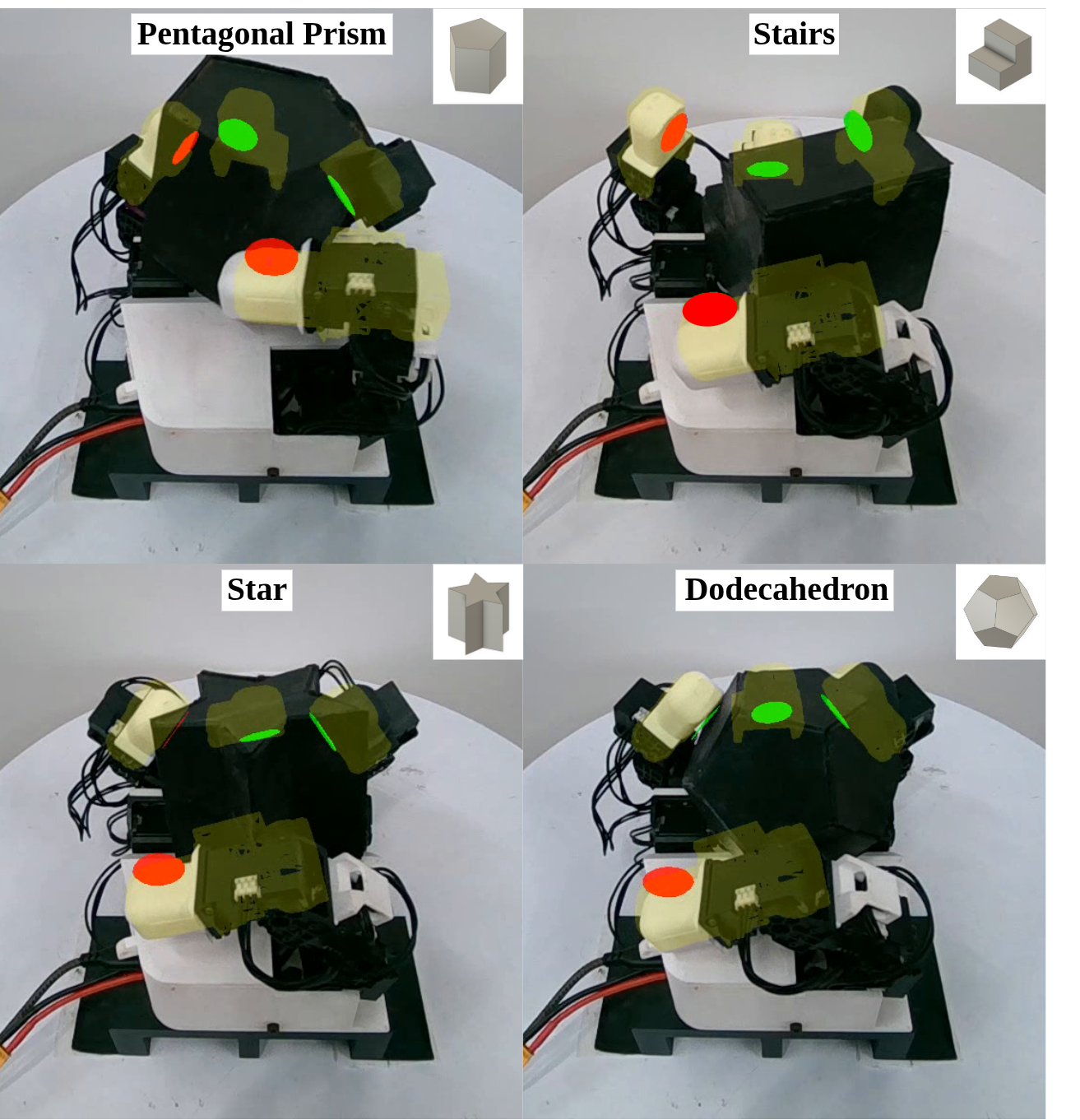}
    \caption{\textbf{Real-World Experimental Evaluation.} Examples of objects being used for downstream policy testing with predicted contacts.}
    \label{fig:hardware_exp}
\vspace{-6mm}
\end{figure}

We next evaluate whether predicted contact signals can effectively substitute for true tactile sensing in closed-loop control, with example setups as depicted in Fig.~\ref{fig:hardware_exp}. We retrain independent policies in simulation with access to ground-truth tactile input and execute them with model-predicted contacts. Performance is assessed over 30-second trajectories using two task-level metrics~\cite{yin2023rotating}: Cumulative Rotation Reward (CRR), derived from the object’s angular velocity about its vertical axis (in simulation), and Cumulative Rotation Angle (CRA), measured in radians (in the real world, manually annotated via video playback by a single human observer, following the standard evaluation protocol of~\cite{yin2023rotating}). All metrics are averaged over 10 independent trials.

Table~\ref{tab:ablation_results} shows that our predicted contacts support closed-loop rotation strongly in both domains, but with a clear sim-to-real gap: performance is uniformly higher in simulation and degrades on hardware, most notably for non-convex geometries (star and letter R) where contact events are partially occluded and evolve rapidly.

Notably, our method achieves higher returns than the hardware Oracle on convex objects, highlighting the sensor sim-to-real gap. Because the policy expects idealized simulator dynamics, real FSRs introduce unmodeled noise that degrades performance. Predicting simulated labels provides an idealized input for simulation-trained policies, buffering against hardware noise when visual occlusion is low. Conversely, on highly occluded non-convex geometries, visual ambiguity dominates, allowing the Oracle to regain the advantage.

The component trends persist across domains but compress in the real world. Pose-only variants with query asymmetry continue to outperform symmetric or MLP fusions. However, the absolute gains over other pose-only baselines shrink on hardware. Removing vision leaves a persistent deficit, demonstrating that proprioception and motion intent alone do not resolve contact ambiguity during control. Vision-only control is the least reliable, mirroring its low-recall contact estimates.

Overall, the downstream results mirror the contact-prediction findings and corroborate our claims: accurate, transferable pseudo-tactile sensing hinges on multimodal fusion with pose-conditioned cross-attention and causal temporal reasoning. These components mitigate (but do not eliminate) the sim-to-real gap; with them in place, predicted contacts provide a practical substitute for hardware tactile signals in dexterous manipulation.
\vspace{-1mm}
\begin{remark}
    The proposed model achieves an average inference latency of \textbf{8 ms} on our workstation after compilation with \texttt{torch.jit} (FP32). This is well below the 50\, ms budget imposed by the 20\, Hz control loop, confirming its suitability for real-time closed-loop manipulation tasks.
\end{remark}

\section{Conclusion, Limitations, and Future Work}
\vspace{-1mm}
\label{sec:conclusion}
We presented a method for predicting binary fingertip contact signals from RGB-D and proprioceptive inputs, providing a focused surrogate for binary tactile sensing in an in-hand object rotation task. Our approach leverages pose-conditioned cross-attention and temporal modeling to resolve visual ambiguities and align these cues with motion intent. We validated the approach extensively in both simulation and on physical hardware. 
The model generalized robustly to unseen objects and held-out policies, including non-convex and irregular geometries. Furthermore, when its predicted contacts were used in place of true tactile feedback, a manipulation policy trained with oracle tactile inputs achieved near-oracle performance without retraining.
These results highlight the practical utility of pseudo-tactile sensing. By providing contact information in the absence of tactile hardware, our method allows policies to retain much of the benefit of tactile feedback, enabling contact-aware control for in-hand manipulation in scenarios where tactile instrumentation is unavailable, unreliable, or impractical.

While our results demonstrate that visuo-proprioceptive contact prediction can serve as a practical surrogate for tactile sensing, several limitations remain. First, the present study focuses on fingertip contacts at a fixed set of sensing sites; extending this framework to denser contact maps and palm regions would further increase its utility. Second, our model is trained on data collected for in-hand dexterous manipulation and may not function as a ``general-purpose'' contact detector for entirely different tasks such as tool use or assembly. A promising path to address this is to develop a foundational model for pseudo-tactile sensing by pre-training on a large and diverse dataset of general hand-object interactions. In parallel, while our frozen asymmetric RGB-D encoder proved highly effective, benchmarking alternative vision backbones remains a key direction for future architectural exploration. Finally, we predict binary contact states and do not estimate richer tactile quantities such as force magnitudes, shear, or torque~\cite{chen2025transforce}. Such signals are crucial in settings involving deformable objects or force-sensitive tasks, and extending pseudo-tactile sensing toward these regimes represents an exciting future direction.
\vspace{-2mm}

\section*{Acknowledgement}
\vspace{-1mm}
\label{sec:acknowledgement}
The authors acknowledge the use of large language models, including Gemini, ChatGPT, and Claude, exclusively for grammatical refinement and enhancement of the manuscript’s linguistic clarity and readability.
\vspace{-2mm}
\bibliographystyle{IEEEtran}
\bibliography{our_bib}
\end{document}